\documentclass[runningheads]{llncs}
\usepackage{graphicx}
\usepackage{multirow}
\usepackage{marvosym}
\usepackage{array}
\usepackage{booktabs}
\usepackage{amsfonts}
\usepackage{float}
\usepackage{amsmath}
\usepackage{CJKutf8}
\usepackage{lipsum}
\usepackage[bookmarks=false,breaklinks,colorlinks,linkcolor=blue,citecolor=blue,urlcolor=black]{hyperref}
\usepackage{color}
\usepackage{stfloats}
\usepackage[linesnumbered,ruled]{algorithm2e}
\usepackage{CJKulem}

\begin{document}
\begin{CJK}{UTF8}{gbsn}
\title{CLTS+: A New Chinese Long Text Summarization Dataset with Abstractive Summaries}

\author{Xiaojun Liu\inst{1,2} \and
Shunan Zang\inst{1,2} \and
Chuang Zhang\inst{1} \textsuperscript{(\Letter)}   \and
Xiaojun Chen\inst{1}\and
Yangyang Ding\inst{1}
}

\authorrunning{X. Liu et al.}
\institute{Institute of Information Engineering, Chinese Academy of Sciences, Beijing, China 
\email{\{liuxiaojun,zangshunan,zhangchuang,chenxiaojun,dingyangyang\}@iie.ac.cn}\and
School of Cyber Security, University of Chinese Academy of Sciences, Beijing, China}

\maketitle              % typeset the header of the contribution
\begin{abstract}
The abstractive methods lack of creative ability is particularly a problem in automatic text summarization. The summaries generated by models are mostly extracted from the source articles. One of the main causes for this problem is the lack of dataset with \normalem \emph{abstractiveness}, especially for Chinese. In order to solve this problem, we paraphrase the reference summaries in CLTS, the \textbf{C}hinese \textbf{L}ong \textbf{T}ext \textbf{S}ummarization dataset, correct errors of factual inconsistencies, and propose the first Chinese Long Text Summarization dataset with a high level of \normalem \emph{abstractiveness}, CLTS+, which contains more than 180K article-summary pairs and is available online\footnote[1]{ \url{https://github.com/lxj5957/CLTS-plus-Dataset}}. Additionally, we introduce an intrinsic metric based on co-occurrence words to evaluate the dataset we constructed. We analyze the extraction strategies used in CLTS+ summaries against other datasets to quantify the \normalem \emph{abstractiveness} and difficulty of our new data and train several baselines on CLTS+ to verify the utility of it for improving the creative ability of models.


\end{abstract}

\section{Introduction}
In the process of human writing summaries, some words, phrases, and sentences that are not in the source article can be produced, which is one of the main differences between human-written and computer-generated summaries. Therefore, in order to make summaries generated by computers as close as possible to the results of human-written, we hope that the model has creative ability, which means that it can generate some novel n-grams, rather than completely copy from the articles, especially for abstractive summarization.

In recent years, deep learning models have greatly improved the state-of-the-art for various supervised NLP problems \cite{vaswani2017attention,paulus2017deep,devlin-etal-2019-bert}, including automatic text summarization \cite{nallapati2017summarunner,see-etal-2017-get,cohan-etal-2018-discourse}. However, the sequence-to-sequence architecture based on deep neural networks is data-driven, which means that the quality and nature of the dataset have a great impact on training and testing the model. Therefore, a high-quality dataset can greatly improve the quality of the summaries generated.

However, there is no \normalem \emph{abstractiveness} dataset in Chinese long text summarization: abstractive summaries describe the contents of articles primarily using new sentences. There is only one Chinese long text summarization dataset called CLTS \cite{liu2020clts}. It is an \normalem \emph{extractiveness} dataset: extractive summaries frequently borrow words and phrases from their source text, which leads to the fact that models trained on CLTS will extract whole sentences from articles to form summaries when predicting. It makes a nonsense of abstractive and reduces the novelty of summaries generated.

In this work, we propose CLTS+ dataset, which is based on CLTS, the \textbf{C}hinese \textbf{L}ong \textbf{T}ext \textbf{S}ummarization dataset. We paraphrase the reference summaries in CLTS to reduce the number of samples that reference summaries are completely extracted from the source articles and make the dataset abstractive to improve the creative ability of models. Meanwhile, some inconsistencies will inevitably occur during the process of paraphrasing; for example, people and place names in summaries after paraphrasing can't be aligned with those in CLTS reference summaries. Therefore, we correct errors of factual inconsistencies to reduce the noise in the dataset and improve the prediction accuracy of models. 

For data-driven learning-based methods (e.g., neural networks), the high quality of the training data ensures that models learn to perform a given task correctly. Therefore, we introduce an intrinsic metric based on co-occurrence words as a supplement to the existing indicators of the dataset quality. This metric focuses on the semantic dimension, and we apply it to CNN/DM, the most commonly used dataset in this field, to verify the reasonability of the metric.

We summarize our contributions as follows: (1)We propose a Chinese long text summarization dataset called CLTS+. To the best of our knowledge, it is the first Chinese long text summarization dataset with such a high level of \normalem \emph{abstractiveness}. There are no article-summary pairs where the reference summary is completely extracted from the source article. (2)We provide an extensive analysis of the properties of this dataset to quantify the \normalem \emph{abstractiveness} and difficulty of our new data. (3)We introduce an intrinsic metric based on co-occurrence words as a supplementary indicator to existing evaluation metrics on dataset. (4)We train and evaluate several baselines on CLTS+ dataset and test on the out-of-domain data to verify the dataset's utility for improving the creative ability of models.

\section{Related Work}
Recently, neural networks have shown great promise in text summarization, with both extractive \cite{zhou-etal-2018-neural,zhong-etal-2019-closer} and abstractive \cite{rush-etal-2015-neural,nallapati-etal-2016-abstractive,see-etal-2017-get,cohan-etal-2018-discourse} methods. End-to-end deep neural models are data-driven, and therefore, there is strong demand for high-quality and large-scale datasets.

However, Chinese datasets for automatic text summarization are only NLPCC, CLTS \cite{liu2020clts} and LCSTS \cite{hu-etal-2015-lcsts}. The size of NLPCC is so small that it can't be used to train neural networks. Take NLPCC-2017 as an example; it only contains 50K article-summary pairs, which is mostly used for evaluation. LCSTS is a short text dataset collected from verified accounts on the Chinese microblogging website Sina Weibo, in which each article doesn't exceed 140 Chinese characters. It is only applicable to abstractive summarization and can only be used to generate a title from a short text. And CLTS, the only Chinese long text summarization dataset, is extracted from the source article, which leads to the fact that models trained on CLTS lose the ability to generate novel n-grams in prediction.

The current situation of English datasets for automatic text summarization is better than Chinese, not only in the number of datasets, but also in the types of them. Commonly used datasets in English include short text Gigaword \cite{rush-etal-2015-neural,chopra-etal-2016-abstractive} and long text CNN/DM \cite{hermann2015teaching,nallapati2017summarunner,see-etal-2017-get}. There are also multi-document datasets including DUC-2004 \cite{over2004introduction}, TAC-2011 \cite{owczarzak2011overview} and Multi-News \cite{fabbri-etal-2019-multi}. In addition to news articles, researchers also introduce dialogue summarization corpus \cite{gliwa-etal-2019-samsum}, patent documents \cite{sharma-etal-2019-bigpatent} and scientific papers \cite{yasunaga2019scisummnet}.

\section{Dataset Construction}
\begin{figure*}
\centering
\includegraphics[scale=0.148]{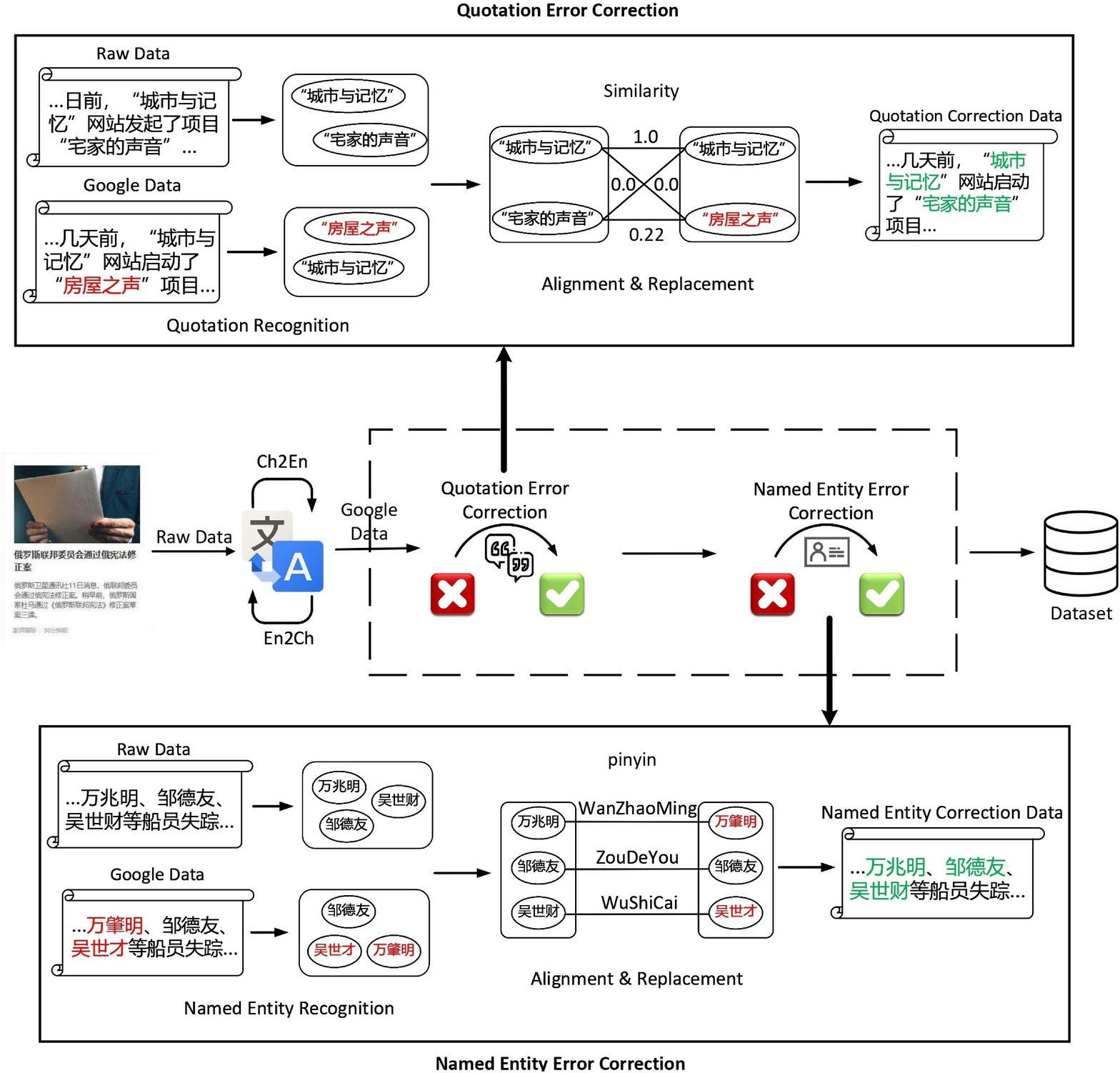}
\caption{Process of CLTS+ dataset construction. \emph{Raw Data} refers to summaries in CLTS. \emph{Google Data} refers to \emph{Raw Data} after back translation. \emph{Quotation Correction Data} and \emph{Named Entity Correction Data} refers to \emph{Google Data} after correcting quotation and named entity errors respectively.}
\label{fig1}
\end{figure*}

The process of constructing CLTS+ dataset is shown in Figure~\ref{fig1}. We leverage back translation to paraphrase reference summaries in CLTS dataset and correct factual inconsistencies such as people and place names. After the operations mentioned above, we obtain the reference summaries in CLTS+ dataset.

\subsection{Back Translation}
Back translation is a commonly used technique for text augmentation in Natural Language Process. It is equivalent to paraphrase the text, which means the semantic meaning remains the same while the expression of the text is completely different from the original. The key idea of back translation is very simple. We translate the original text from the source language into another language and then translate it back into the source language.

There are many machine translation services to translate to a different language and back to Chinese, such as Google Translate, Baidu Translate, etc. Google Translate is the most popular service among them. However, it is needed to get an API key to use it, and it is a paid service. Alternatively, we use a handy feature named GOOGLETRANSLATE() in Google Sheets\footnote[2]{\url{https://www.google.cn/intl/zh_cn/sheets/about/}} web app to leverage for our purpose. We translate the reference summaries in CLTS dataset from Chinese to English and then back again to Chinese. After the back translation process, we have achieved the purpose of paraphrasing the reference summaries in CLTS dataset, which formed by extractive fragments from the source articles have transformed into more abstractive ones.

\subsection{Error Correction}
Since most of the samples in the dataset are news articles, the reference summaries will contain the quotation of statements in the source articles, which includes but is not limited to important details of the speeches delivered by national leaders and the views of the interviewees on an event. These statements are also objective facts, and therefore, although the semantic meaning before and after back translation is the same, we still hope that the quotation of such statements should not be paraphrased.

We recognize quotations with regular expression in Raw Data $\boldsymbol{R}$ and Google Data $\boldsymbol{G}$, and then compute the similarity, which is defined by the normalized length of longest contiguous matching subsequence, between quotations in $\boldsymbol{R}$ and $\boldsymbol{G}$. Two quotations with a similarity higher than the threshold are regarded as aligned, and we replace the quotation in Google Data $\boldsymbol{G}$ with the one in Raw Data $\boldsymbol{R}$.

Additionally, during the process of back translation, errors of named entities will inevitably occur, among which the commonest are incorrect factual details such as people and place names. If these errors are not corrected, samples containing them are noise in the dataset, which will negatively affect model training and reduce the accuracy when inferring. Therefore, the errors of named entities need to be corrected so that models can predict accurately during testing.

We use jieba\footnote[3]{\url{https://github.com/fxsjy/jieba}} to make word segmentation on Raw Data $\boldsymbol{R}$ and Google Data $\boldsymbol{G}$ respectively. We determine whether a word is a named entity based on surname and the length of it. After named entity recognition, we align named entities in $\boldsymbol{R}$ and $\boldsymbol{G}$ by pinyin, because as for people's name, the pinyin is always the same, although Chinese characters are different during the process of back-translation. Finally, we replace named entities in $\boldsymbol{G}$ with those in $\boldsymbol{R}$.

We correct 42,030 quotation errors and 263,822 named entity errors. After back translation and error correction, we offered CLTS+ dataset containing 181,401 samples.

\subsection{Metric Based on Co-occurrence Words}
High-quality data forms the bedrock for building meaningful statistical models in NLP. In order to ensure the quality of the dataset we constructed, we introduce an intrinsic metric based on co-occurrence words as a supplement to the existing indicators. The process is shown in Figure~\ref{fig2}, and the main steps are summarized as follows:

\begin{figure*}
\centering
\includegraphics[scale=0.0845]{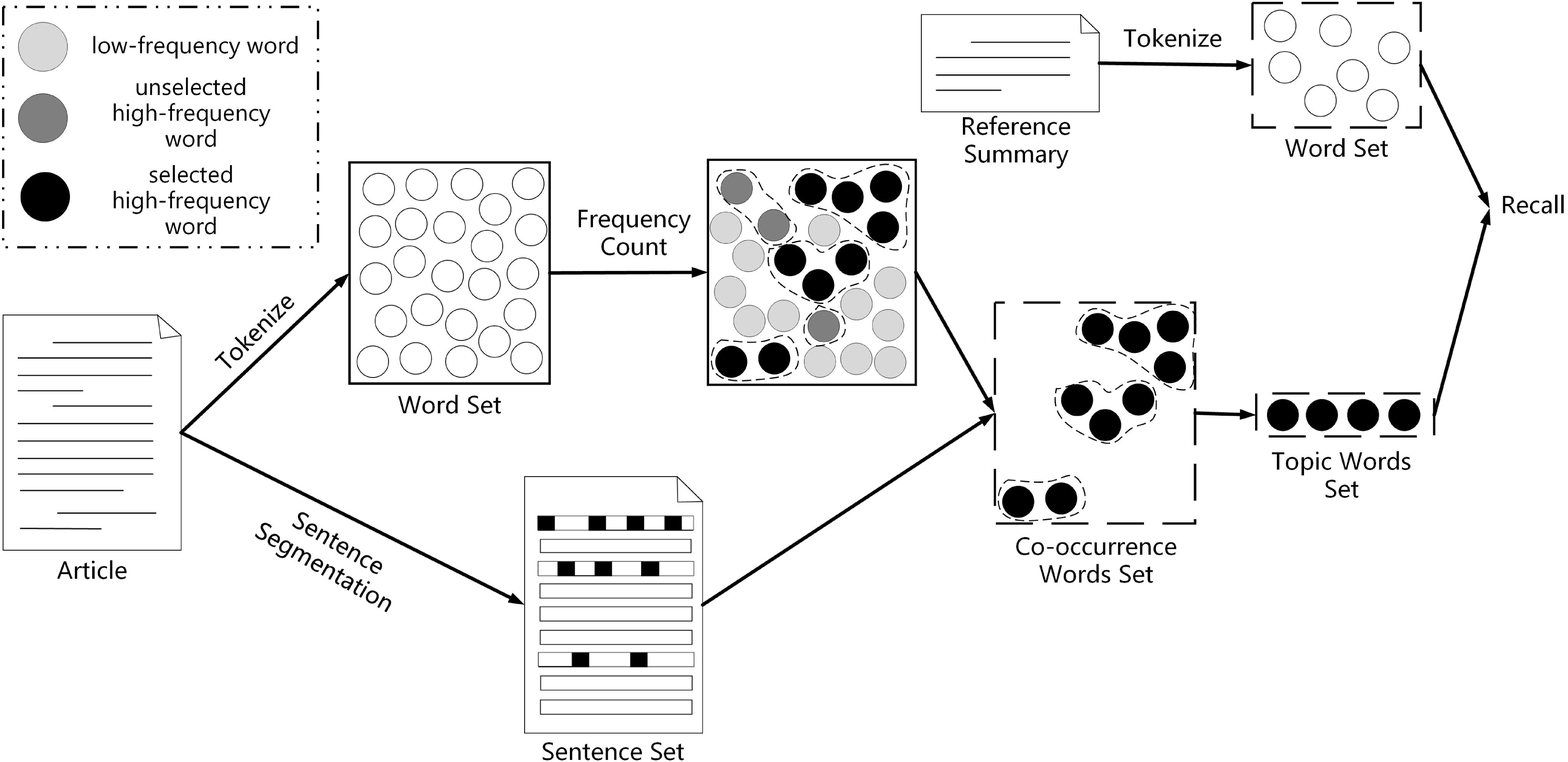}
\caption{Process of our metric based on co-occurrence words.}
\label{fig2}
\end{figure*}

(1)	Tokenize the article and remove stop words from it to get the Word set $\mathcal{W}$.

(2)	Sort $\mathcal{W}$ in descending order according to the word and sentence frequency. Sentence frequency refers to a sentence that contains the word occurring how many times in the article. And then take the first N words to obtain the High-frequency Word set $\mathcal{HW}$.

(3)	Perform sentence segmentation on the article to get the Sentence set $\mathcal{S}$. We define co-occurrence words as high-frequency words in $\mathcal{HW}$ appear in the same sentence. Sort the co-occurrence words in descending order according to the length and frequency of them to obtain the co-occurring word set and take the top-K co-occurrence words as Topic Word set $\mathcal{TW}$. We regard $\mathcal{TW}$ as a highly condensed expression of the article.

(4)	Tokenize the reference summary and compute recall of the words in reference summary in $\mathcal{TW}$. This metric assigns a value $x\in[0, 1]$ to every document-summary pairs $(D_i, S_i)$ where 1 is the maximum score and example-level scores are averaged to yield a dataset-level score.

\section{Analysis}
\subsection{Data Statistics}

\begin{table*}
\caption{An example of CLTS+ and CLTS dataset. The red font means extractive fragments of an article-summary pair.}
\label{tab1}
\centering
\resizebox{\textwidth}{!}{
\begin{tabular}{c|m{111mm}}
\hline
\rule{0pt}{54pt} \textbf{Article(truncated)} & …\textcolor[rgb]{1,0,0}{工信部党组成员、总工程师}张峰出席博览会并致辞。他在会上透露，接下来将着力完善三大体系，构筑工业互联网生态。…\textcolor[rgb]{1,0,0}{张峰提出，我国工业互联网应用场景丰富，模式创新活跃，但关键核心技术还不足。因此，要加强产业创新能力，夯实持续发展基础。}…(… Zhang Feng, a member of the Party Leadership Group and Chief Engineer of the MIIT, attended the Expo and delivered a speech. He said at the meeting that three systems would be improved, and an industrial Internet ecosystem would be built in follow-up work. … Zhang Feng pointed out that there are rich application scenarios and active innovation patterns in the industrial Internet in China. However, the key technologies are still insufficient. Therefore, it is necessary to strengthen industrial innovation capabilities and consolidate the foundation for sustainable development. … )\\
\hline
\rule{0pt}{14pt} \textbf{Summary in CLTS}  &  \hspace*{1em} \textcolor[rgb]{1,0,0}{工信部党组成员、总工程师张峰提出，我国工业互联网应用场景丰富，模式创新活跃，但关键核心技术还不足。因此，要加强产业创新能力，夯实持续发展基础。}    \\
\hline
\rule{0pt}{14pt} \textbf{Summary in CLTS+}  &  \hspace*{1em} 工业和信息化部\textcolor[rgb]{1,0,0}{党组成员，总工程师张峰}指出，我国的工业互联网具有丰富的应用场景和积极的模式创新，但关键的核心技术仍然不足。因此，有必要\textcolor[rgb]{1,0,0}{加强产业创新能力}，夯实可持续发展基础。  \\
\hline
\end{tabular}
}
\end{table*}

\begin{table*}
\caption{Dataset Statistics. For Chinese dataset, length is the number of Chinese characters, while for English, it is the number of words.}
\label{tab2}
\centering
\resizebox{\textwidth}{!}{
\begin{tabular}{c|c|cc|cc|cc}
\hline
\multirow{2}{*}{Dataset} & \multirow{2}{*}{\#docs(train/val/test)} & \multicolumn{2}{c|}{avg.document length} & \multicolumn{2}{c|}{avg.summary length} & \multicolumn{2}{c}{vocabulary size} \\
                         &                                         & words             & sentences            & words            & sentences            & douments         & summeries        \\ \hline
CLTS+ &  145,118 / 19,954 / 16,327  &  1,584.76  &  32.53  &   68.79  &  1.95  &  215,553  &  23,773  \\
CLTS  &  148,317 / 20,393 / 16,687  &  1,582.88  &  32.51  &   66.34  &  1.50  &  220,428  &  18,981  \\
CNN/DM&  287,227 / 13,368 / 11,490  &  791.36    &  42.65  &   62.75  &  4.85  &  722,715  &  197,771   \\\hline
\end{tabular}}
\end{table*}

CLTS+ dataset contains 181,401 article-summary pairs, and is divided into training (145,120 samples, 80\%), validation (19,954 samples, 11\%) and testing (16,327 samples, 9\%) sets. Table~\ref{tab1} shows the same article corresponding to different reference summaries in CLTS+ and CLTS respectively, which indicates that the reference summaries in CLTS+ are quite abstractive. Simple statistics on CLTS+ dataset compared to other datasets are shown in Table~\ref{tab2}. To quantify \normalem \emph{abstractiveness} in our dataset, we use one measure named novel n-grams, which refers to unique n-grams in the summary which are not in the article. The results are shown in Table~\ref{tab3}.

\subsection{Rouge of LEAD-3 And Oracle}
LEAD-3, which selects the first three sentences of the article, is a strong baseline in automatic summarization. News articles adhere to a writing structure known in journalism as the “Inverted Pyramid.” In this form, the initial paragraphs contain the most newsworthy information, which is followed by details and background. This structure makes LEAD-3 can be competitive with some state-of-art systems, although it is very simple. 

Oracle represents the best possible performance of an ideal extractive system. Given an article $A$ and its summary $S$, Oracle summary refers to concatenating the fragments in $\mathcal{F}(A, S)$ \cite{grusky-etal-2018-newsroom} in the order they appear in the summary $S$.

We report ROUGE \cite{lin-2004-rouge} scores, which measures the overlap of unigrams (ROUGE-1), bigrams (ROUGE-2), and longest common subsequence (ROUGE-L) between the candidate and reference summaries. If summaries generated by LEAD-3 and Oracle achieve a high ROUGE score, it means that the reference summaries are directly extracted from the source articles with no abstractiveness. In that case, the dataset can’t be used to challenge the summarization task and verify the effectiveness of models because the nature of the dataset determines that the most sophisticated models can’t surpass the simple, strong baseline LEAD-3 and Oracle.

\begin{table}
\caption{ROUGE scores for LEAD-3 and Oracle on CLTS+, CLTS and a commonly used English dataset, CNN/DM.}
\label{tab3}
\centering
\begin{tabular}{c|cccc|ccc|ccc}
\hline
\multirow{2}{*}{Dataset} & \multicolumn{4}{c|}{\% of novel n-grams in gold summary} & \multicolumn{3}{c|}{LEAD-3} & \multicolumn{3}{c}{ORACLE} \\
                         & unigrams      & bigrams     & trigrams     & 4-grams     & R-1     & R-2     & R-L     & R-1     & R-2     & R-L    \\ \hline
CLTS+  & 8.66\% & 38.07\% & 55.77\% & 66.37\% & 42.27 & 22.51 & 35.34 & 83.05  & 71.03 & 84.01  \\
CLTS   & 0.57\% & 3.23\%  & 5.38\%  & 7.10\%  & 49.87 & 37.78 & 45.73 & 99.50  & 99.08 & 99.50  \\
CNN/DM & 11.10\%& 48.71\% & 69.51\% & 79.53\% & 40.34 & 17.70 & 36.57 & 71.49  & 56.15 & 76.44   \\\hline
\end{tabular}
\end{table}

Table~\ref{tab3} presents the ROUGE score of summaries predicted by LEAD-3 and Oracle on CLTS+, CLTS, and CNN/DM test sets. CLTS dataset has achieved the highest ROUGE score, whether on LEAD-3 or Oracle, which shows that the reference summaries in CLTS are all extracted from the source articles. The Oracle ROUGE score of CLTS+ is higher than CNN/DM. In comparison, the LEAD-3 score is lower than it and drops significantly compared with CLTS, which indicates that although the reference summaries in CLTS+ copy some words from the source articles, the reorganization of them also makes the summaries abstractive.

\subsection{Human Evaluation}
The metrics mentioned above are all automatic. In order to further verify the quality of CLTS+, we provide human evaluation of the different datasets on the reference summaries and articles. Human evaluation is centered around five dimensions, including fluency (individual sentences are well-written and grammatical), coherence (all sentences of the summary fit together and make sense collectively), consistency (factual details in the summary are aligned with the ones in the article), informativeness (summary captures the key points from the article) and novelty (words, phrases, and sentences of the summary are not in the article).

Evaluation is performed on 100 randomly sampled articles with the system-generated summaries of Pointer-Generator network. Each system-article pair is rated by five distinct judges, with the final score obtained by averaging the individual scores (out of 5). 

\begin{table*}
\caption{Human evaluation on the system-generated summaries of Pointer-Generator network trained on CLTS, CLTS+ and CNN/DM respectively.}
\label{tab4}
\centering
\begin{tabular}{c|ccccc}
\hline
Dataset  & Fluency       & Coherence     & Consistency     & Informativeness & Novelty          \\ \hline
CLTS     &\textbf{3.92}  & \textbf{3.87} &   \textbf{3.90} &    2.09         &  0.03            \\
CLTS+    &  3.71         &   3.60        &   3.87          &    3.40         &  \textbf{3.80}   \\
CNN / DM &  3.56         &   3.39        &   3.62          & \textbf{3.55}   &  3.73            \\ \hline
\end{tabular}
\end{table*}

Table~\ref{tab4} shows human evaluation results on CLTS, CLTS+ and CNN/DM. Reference summaries in CLTS get the highest fluency and coherence scores because they are completely extracted from the source articles. Although it ensures the correctness of syntactic, the novelty score is almost zero. CLTS+ exhibits similar to CNN/DM dataset on all scores, and it greatly improves the novelty score without sacrificing fluency and coherence.

\subsection{Our Metric}
When performing the metric we introduced on different datasets, We set N=8 in Chinese, N=20 in English, and K=1. The average score on CLTS+, CLTS, and CNN/DM is 0.384, 0.522, and 0.496, respectively. The percentage of samples in the three datasets mentioned above get a score between (0.5, 1.0] is 27.27\%, 46.21\%, 44.48\%. Compared with CNN/DM, the most commonly used dataset in automatic text summarization, CLTS+ achieves a similar distribution and high consistency with it, which demonstrates the high quality of the dataset we constructed and shows the reasonability of the metric.

% \begin{table}[]
% \caption{Results on different datasets evaluated by our metric.}
% \label{tab5}
% \centering
% \begin{tabular}{c|c|c|c}
% \hline
%            & CLTS+   & CLTS & CNN/DM \\ \hline
% Average    & 0.384   &  0.522    & 0.496 \\ \hline
% (0.5, 0.8] & 20.6\%  & 24.0\%    & 33.9\%  \\
% (0.8, 0.9] &  4.14\% & 8.51\%    & 7.19\%       \\
% (0.9, 1.0] &  2.53\% & 13.7\%    &  3.39\%      \\ \hline
% \end{tabular}
% \end{table}

\section{Experiment}
\subsection{Baseline}
We train and evaluate several summarization systems to understand the challenges of CLTS+ and its usefulness for training systems. We consider two unsupervised extractive systems including LEAD-3 and TextRank \cite{mihalcea-tarau-2004-textrank}. Additionally, we train four abstractive systems: Seq2seq with attention (Seq2seq-att) \cite{chopra-etal-2016-abstractive,nallapati-etal-2016-abstractive}, Pointer-Generator (Pointer-gen) and a version with coverage mechanism (Pointer-gen+cov) \cite{see-etal-2017-get} and Transformer \cite{vaswani2017attention}.

\subsection{Automatic Evaluation}
For abstractive models mentioned above, we take the Chinese character as input and use OpenNMT-py \cite{klein-etal-2017-opennmt} to re-implement them. The results are shown in Table~\ref{tab6}.

All baselines perform better on CLTS+ compared to CNN/DM. Although ROUGE scores on CLTS+ are lower than CLTS, they are incomparable because their extractive strategies are different. The results on CLTS+ are consistent with CNN/DM, the most commonly used English dataset, which means that CLTS+ can be used as a benchmark dataset in automatic text summarization.

\begin{table*}
\caption{ROUGE scores of different methods on CLTS, CLTS+ and non-anonymized version of CNN/DM dataset.}
\label{tab6}
\centering
\begin{tabular}{c|ccc|ccc|ccc}
\hline
                & \multicolumn{3}{c|}{CLTS} & \multicolumn{3}{c|}{CLTS+} & \multicolumn{3}{c}{CNN/DM} \\ \hline
Model           & R-1     & R-2    & R-L    & R-1     & R-2     & R-L    & R-1      & R-2     & R-L     \\ \hline
LEAD-3          & 43.18   & 35.79  & 40.54  & 37.01   & 20.89   & 34.76  & \textbf{40.34}    & 17.70   & 36.57   \\
TextRank        & 31.55   & 22.50  & 27.00  & 27.45   & 13.90   & 25.38  & 35.23    & 13.90   & 31.48   \\ \hline
Seq2seq-att     & 48.05   & 37.00  & 43.34  & 41.35   & 24.03   & 36.47  & 31.33    & 11.81   & 28.83   \\
Pointer-gen     & \textbf{51.68}   & \textbf{41.17}  & \textbf{47.18}  & 42.06   & 24.90   & 37.11  & 36.44    & 15.66   & 33.42   \\
Pointer-gen+cov & 46.34   & 34.88  & 41.39  & 40.65   & 23.21   & 35.90  & 39.53    & 17.28   & 36.38   \\
Transformer     & 48.92   & 37.91  & 44.27  & \textbf{43.27}   & \textbf{26.17}   & \textbf{37.97}  & 40.05    & \textbf{17.72}   & \textbf{36.77}   \\ \hline
\end{tabular}
\end{table*}

\subsection{Results on Out-of-domain Data}
In addition to generating summaries on the test set, we also evaluate Pointer-Generator network trained on CLTS and CLTS+ respectively on the out-of-domain data to verify the creative ability of the model. We randomly select 100 news articles from Chinese news website such as ChinaNews\footnote[4]{\url{https://www.chinanews.com/}}, TencentNews\footnote[5]{\url{https://news.qq.com/}} and SohuNews\footnote[6]{\url{http://news.sohu.com/}}.

Table~\ref{tab7} is an example of computer-generated summaries of the models trained on CLTS and CLTS+ respectively, and the article is from ChianNews\footnote[7]{\url{http://www.bj.chinanews.com/news/2020/1020/79344.html}}. The summary generated by the model trained on CLTS+ is not only abstractive but also informative, which includes the information "北京市政府新闻发言人" that is not in the system-generated summary of the model trained on CLTS.

\begin{table*}
\centering
\caption{An example on the out-of-domain data of Pointer-Generator model trained on CLTS and CLTS+ respectively. The underlined words are extractive fragments of an article-summary pair.}
\label{tab7}
\resizebox{\textwidth}{!}{
\begin{tabular}{c|m{90mm}}
\hline
\rule{0pt}{41pt} \textbf{Article(truncated)} & \hspace*{1em}中新社北京10月20日电 (记者 于立霄)\uline{北京市政府新闻发言人}徐和建在20日举行的疫情防控发布会上表示，要继续做好直航北京国际航班的管控，严格防疫措施和闭环管理；严防第三国人员中转入境进京。

\hspace*{1em} \uline{徐和建指出，北京要密切关注重点地区疫情走势，对中高风险地区进京人员迅速排查、加强管控，落实核酸检测、医学观察等措施。}同时，要加强进口冷链食品监管，严格口岸查控措施，督促承运企业落实主体责任，做好从业人员防护和运输装备、外包装消毒。…\\
\hline
\rule{0pt}{15pt} \textbf{Model trained on CLTS}  &  \hspace*{1em} \uline{徐和建指出，北京要密切关注重点地区疫情走势，对中高风险地区进京人员迅速排查、加强管控，落实核酸检测、医学观察等措施。} \\
\hline
\rule{0pt}{15pt} \textbf{Model trained on CLTS+}  &  \hspace*{1em} \uline{北京市政府新闻发言人} \hspace*{0.3em} \uline{徐和建指出}，\uline{北京}必须\uline{密切关注重点地区}的流行病，\uline{对中高风险地区}进行\uline{迅速}调查，\uline{加强}管理控制，实施\uline{核酸检测、医学观察等措施。} \\
\hline
\end{tabular}
}
\end{table*}

\section{Conclusion}
We present CLTS+ dataset, the largest Chinese long text summarization dataset with a high level of \normalem \emph{abstractiveness}, consisting of various articles and corresponding abstractive summaries. The results show that the dataset can improve the creative ability of models and be used as a benchmark. We hope that the new dataset will promote the development of automatic text summarization research as a new option for researchers to evaluate their systems.

\bibliographystyle{splncs04}
\bibliography{anthology,custom}

\end{CJK}
\end{document}